# A sparse annotation strategy based on attention-guided active learning for 3D medical image segmentation

Zhenxi Zhang[1(✉)], Jie Li[1], Zhusi Zhong[1], Zhicheng Jiao[2] and Xinbo Gao[1]

[1] School of Electronic Engineering, Xidian University, Xi'an 710071, China
xidiangulasijia@outlook.com
[2] Department of Radiology and BRIC, University of North Carolina at Chapel Hill, Chapel Hill, NC 27599, USA

**Abstract.** 3D image segmentation is one of the most important and ubiquitous problems in medical image processing. It provides detailed quantitative analysis for accurate disease diagnosis, abnormal detection, and classification. Currently deep learning algorithms are widely used in medical image segmentation, most algorithms trained models with full annotated datasets. However, obtaining medical image datasets is very difficult and expensive, and full annotation of 3D medical image is a monotonous and time-consuming work. Partially labelling informative slices in 3D images will be a great relief of manual annotation. Sample selection strategies based on active learning have been proposed in the field of 2D image, but few strategies focus on 3D images. In this paper, we propose a sparse annotation strategy based on attention-guided active learning for 3D medical image segmentation. Attention mechanism is used to improve segmentation accuracy and estimate the segmentation accuracy of each slice. The comparative experiments with three different strategies using datasets from the developing human connectome project (dHCP) show that, our strategy only needs 15% to 20% annotated slices in brain extraction task and 30% to 35% annotated slices in tissue segmentation task to achieve comparative results as full annotation.

## 1 Introduction

Medical image segmentation plays an important role in disease diagnosis, pathological analysis and surgical planning. U-net and u-net like models [2, 3, 4] developed from Fully Convolutional Networks [1] have shown good performance in segmenting medical images, and u-net has become the standard structure in medical image segmentation.

Most medical images are 3D volumetric data. 3D segmentation is a voxel-by-voxel classification problem. The classified voxels have spatial relationships with the feature maps (FMs) in the convolution process. However, not all FMs have the same impact for a set of voxels to be classified. The FMs are needed to recalibrate adaptively, to emphasize meaningful features, while weakening irrelevant ones. Many previous works did not emphasize this problem. The "Squeeze-and-Excitation" (SE) block [6] is the application of attention mechanism in convolutional neural networks (CNNs). It recalibrates channel-wise FMs adaptively and achieves better performance in multiple datasets and tasks. Gaining inspiration from SE block, three variants of SE modules were proposed in [7]. All of these attention blocks can be embedded in u-net.

Another enormous challenge for 3D medical image segmentation is that it is very difficult to obtain medical image with full annotation. Only anatomical experts can annotate medical images and it takes a lot of manpower and time to annotate one sample. Sparse annotation is used in many works to reduce annotation time. A small number of 2D annotated slices can generate 3D dense segmentation in 3D u-net [3]. Another 3D instance segmentation model was trained with bounding boxes for all instances and a small number of annotated instances [5].These works didn't provide a selection strategy for sparse annotation.

Accurate selection of the most informative slices to label involves the problem of active learning. Active Learning (AL) has been well researched over the past few decades. AL frameworks improve existing models by incremental selection and annotation of the most informative unlabeled samples, and applied on various vision tasks. The method AIFT (active, incremental fine-tuning) proposed in [8] integrated active learning and transfer learning into a framework and applied on three different biomedical imaging tasks. AIFT starts with pre-trained CNN, and selects informative samples from the unannotated for annotation based on entropy and diversity. A deep active learning framework [9] was proposed to reduce annotation cost, which starts with less training data to train a set of FCNs, then seeks informative samples for annotation in term of the uncertainty estimated from the trained FCNs and the similarity measured between images. The key to active learning is to find a criterion to select the most valuable samples for labeling. The above two works select slices based on the uncertainty and similarity. Although they are related to the real accuracy, they cannot be linked to the final evaluation criteria directly.

As shown in Fig. 3, as the number of training iterations increases, attention weight tends to get larger values in the target area and smaller values in the background. The distribution of attention weight is closer to the distribution of the ground truth, like the visualization of attention maps in [10], which helps to improve image segmentation accuracy. We speculate that attention mechanism may estimate the segmentation accuracy and provide a guidance for active learning. So we embed attention mechanism into 3D u-net and design an attention-guided active learning strategy.

The contributions of this paper are summarized as follows: (1) we propose an accurate and efficient suggestive annotation strategy based on attention-guided active learning for 3D segmentation in medical imaging. (2) The channel attention mechanism (CAM) and the spatial attention mechanism (SAM) are embedded into 3D u-net in a combined way, which not only improves the segmentation accuracy, but also feeds back the useful information of the trained model to the unlabeled. (3) The pseudo dice similarity coefficient (P-DSC) and the pseudo accuracy (P-accuracy) correctly predict the segmentation quality of unlabeled slices and have a great positive correlation with the real segmentation accuracy. (4) We only use 15% to 20% annotated slices for brain extraction and 30% to 35% annotated slices for tissue segmentation to train our proposed model, gaining the comparative results with fully supervised training.

## 2 Method

As shown in Fig. 1, where *I* means input, *O* means output, our approach can be summarized as follows. Attention-embedded 3D u-net is adopted as the segmentation model. The model is trained by initial training set $O_{3\_0}/I_{1\_1}$ firstly. In the i-th iterative training, the weakly supervised training method is used to train the proposed model with slice-wise sparsely labeled training sets ($I_{1\_i}$). The final FMs and its attention maps (AMs) ($O_{1\_i}$) can be outputted by segmentation model. The output of segmentation model ($O_{1\_i}$) serves as the input of the active learning model ($I_{2\_i}$). The accuracy of segmentation results predicted by the current model is estimated by the confidence degree of the AMs to the final FMs. The indexes of the slices to be labeled ($O_{2\_i}$) are exported by the active learning model. The process of sparse labeling is simulated using fully annotated dataset. Then the training set is updated and the segmentation model is fine-tuned with new training set ($O_{3\_i}/I_{1\_[i+1]}$) until the stable and high-quality segmentation performance is obtained.

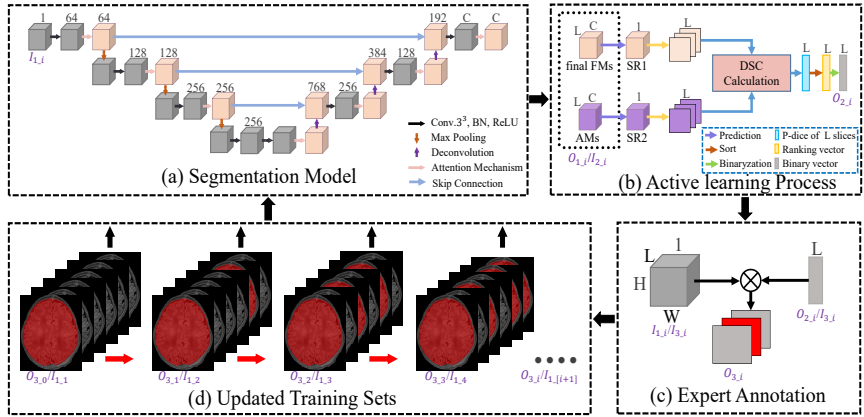

**Fig. 1.** Overall framework of our proposed method: (a) Attention-embedded 3D u-net. (b) Attention-guided active learning process. (c)Expert Annotation simulation. (d) Updated training sets. It should be noted that the final FMs and its AMs become a bridge between (a) and (b).

### 2.1 Attention-embedded 3D u-net

3D u-net is an extension of u-net, which is applied to 3D image segmentation. The architecture of 3D u-net is illustrated in segmentation model part in Fig.1 (a). The left side can be regarded as an encoder and the right side as a decoder. We add attention mechanism to recalibrate the FMs in 3D u-net. The skip connection is used to concatenate the up-sampling results with the same resolution FMs in the encoder as the input of the decoder submodule. Each decoder submodule contains a deconvolution operation, recovering the FMs to the same resolution as the input resolution gradually.

The structure of the attention mechanism is shown in Fig. 2. Convolution can be seen as a process of aggregating spatial information and channel-wise information. The attention mechanism embedded in 3D u-net consists of two parallel paths [16]. The first path is CAM . The purpose of CAM is to automatically acquire the importance of each channel by learning. The second path is SAM, which takes effect on the spatial scale, weighting different regions in the FMs to make the segmentation model focus on the important feature areas, highlighting the salient regions. Finally, the input FMs are re-weighted by multiplying combined AMs. It should be noted that only the AMs of final FMs are returned for active learning.

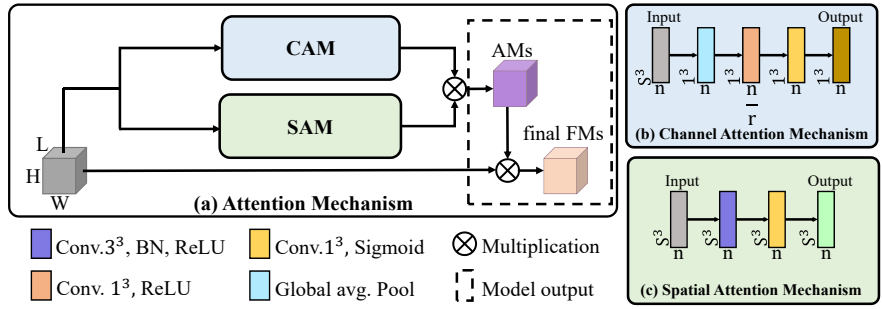

**Fig. 2.** Schematic of the proposed attention mechanism. **(a)** The structure of the attention mechanism we use. **(b)** CAM structure **(c)** SAM structure. Conv. $3^3$ denotes 3×3×3 convolution.

### 2.2 Attention-guided active learning

Uncertainty measure is the core of active learning. It can be estimated by model output. We propose a new measure of uncertainty in segmentation tasks which uses AMs. At the beginning, we use the data annotated every 32 or 16 slices from the axial plane to train our proposed model .The slice similarity in 3D image gradually decreases as the spatial distance of slice increases. So this distance can make the slices we pick out more representative. As shown in active learning process in Fig. 1 (b), the segmentation model output (final FMs and its AMs) is input to active learning model. C is the number of classes and L is the maximum value of the axial index. The segmentation result1 (SR1) and the segmentation result2 (SR2) are predication of the final FMs and its AMs respectively. Then the DSCs are calculated slice by slice on axial view. We call this coefficient P-DSC. Then the slices are sorted by P-DSC, generating an ordering vector. Excluding annotated slices, a certain number of slices with lower P-DSC are selected by the binary vector. For tissue segmentation task which is a multi-class segmentation problem, the number of classes on different slices may be different, and the difficulties in segmentation of different tissues vary due to the diversity in brain tissue morphology. Calculating the average P-DSC of all tissues existing in one slice may not reflect the segmentation accuracy of this slice. Another coefficient is proposed for this case. We calculate the number of voxel points predicated to be identical between $SR1_{[i]}$ and $SR2_{[i]}$ except the background class 0, where i denotes the i-th slice on axial view. Then it is divided by the total number of non-0 voxel points in $SR1_{[i]}$ and $SR2_{[i]}$. We call this coefficient P-accuracy, which is corresponding to the real accuracy (R-accuracy). $GT_{[i]}$ denotes the ground truth of the i-th slice. The formulas of the real evaluation coefficients and the uncertainty measures (P-DSC and P-accuracy) are defined as follows:

$$R - DSC\ of\ slice_{[i]} = \frac{2|SR1_{[i]} \cap GT_{[i]}|}{|SR1_{[i]}| + |GT_{[i]}|} \quad (1)$$

$$P - DSC\ of\ slice_{[i]} = \frac{2|SR1_{[i]} \cap SR2_{[i]}|}{|SR1_{[i]}| + |SR2_{[i]}|} \quad (2)$$

$$R - accuracy\ of\ slice_{[i]} = \frac{2|SR1_{[i]} \cap GT_{[i]}|}{|SR1_{[i]} \neq 0| + |GT_{[i]} \neq 0|} \quad (3)$$

$$P - accuracy\ of\ slice_{[i]} = \frac{2|SR1_{[i]} \cap SR2_{[i]}|}{|SR1_{[i]} \neq 0| + |SR2_{[i]} \neq 0|} \quad (4)$$

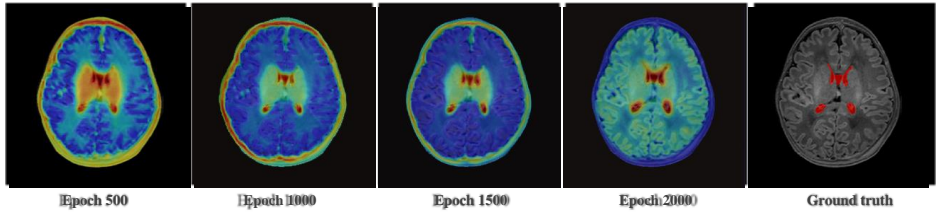

Epoch 500   Epoch 1000   Epoch 1500   Epoch 2000   Ground truth

**Fig. 3.** This figure shows the changes of the AMs' channel about the ventricle as our proposed framework proceeds, which illustrates that AMs can provide a guidance for active learning in segmentation problems.

### 2.3   Expert annotation simulation

In practice, the slices selected by active learning can be authentically labeled by experts. In our experiments, we set the unlabeled slices to another class. For example, in brain extraction task, different samples are given the label 0: "background", 1: "brain", 2: "unlabeled slices". The expert annotation process is simulated by removing the masks from the unlabeled slices and reverting to the original annotation. We cycle on until the average P-DSC or P-accuracy of all slices changes less than a threshold σ. According to a large number of experiment results, it is more suitable to set σ to 0.005.

## 3   Experiments and Results

### 3.1   Training

We evaluated our method on a brain MRI dataset of 40 neonates at term age from the developing Human Connectome Project(dHCP) [12].This dataset provides ground truth of brain extraction (Task1) and tissue segmentation (Task2) [13,14,15]. Slice-wise sparsely labeled training sets are used for training. We adopt the weighted cross-entropy loss and set the weight of the unlabeled class to zero. We implement the proposed method based on Pytorch. Optimization is carried out for 500 iterations every updated training set, and we use the SGD algorithm with learning rate = 0.001. Batch size, momentum and weight decay coefficients are set to 2, 0.9 and 0.0001 respectively. Two NVIDIA GTX1080Ti GPUs with 11GB memory are used for training.

### 3.2 Effective Active learning strategy

In the iterative training process, we randomly select a sample for two tasks and put it into the model for prediction. From the Fig. 4, we can see that the P-DSC /P-accuracy are positively correlated with the R-DSC/R-accuracy. So the segmentation accuracy can be correctly estimated by two metrics, like the metrics proposed in [11], and the most informative slices can be selected.

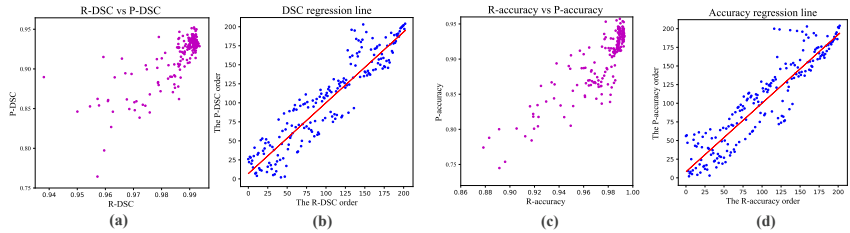

**Fig. 4.** For every slice from the axial view, we calculate the real segmentation accuracy (based on R-DSC or R-accuracy) and estimate the accuracy (based on P-DSC or P-accuracy) respectively. In (b) and (d), let x be the order of real accuracy in all slices, y be the order of accuracy estimation in all slices, and draw scatter plots based on $(x, y)$. Linear regression was carried out with these scatters. The regression coefficient is 0.93 in (b), and 0.92 in (d).

### 3.3 Results

The experimental results of different methods are listed in Table 1, which illustrates that our proposed segmentation model improves the segmentation accuracy. Our proposed model shows good performance (∼ 2.4% better than 3D u-net) in Task2. Fig. 5 shows the effectiveness of our proposed strategy. We compare our strategy with three different annotation strategies: (1) random query: randomly selecting slices to label, (2) equal-interval query: selecting slices at an equal distance, and (3) uncertainty query: selecting slices based on the sum of cross entropy of each voxel point in one slice. In order to compare with equal-interval query, we choose the appropriate annotation ratio to predict, for example, spacing of 8 corresponds to the annotation ratio 12.5%. Meanwhile, to find the minimum labeling cost, we use the trained models in our strategy to predict results once every 5% annotation is added. Two parts of dataset are used to predict. Part A is the sparsely labeled training sets. Part B is the unlabeled test sets.

**Table 1.** Comparison with mainstream methods using full annotated training sets

| Method | F1 score (Task1) | F1 score (Task2) |
|---|---|---|
| 3D U-Net [3] | 0.976 | 0.848 |
| V-Net [4] | 0.975 | 0.844 |
| Our method | 0.979 | 0.872 |

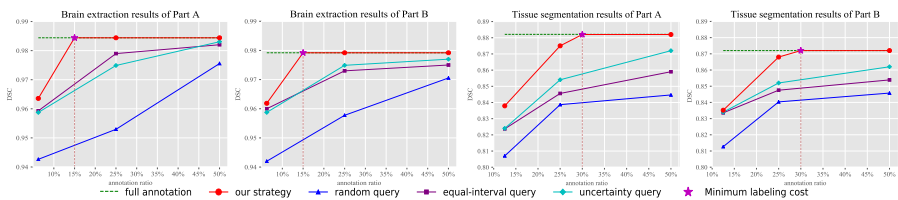

**Fig. 5.** This figure shows comparisons of our approach with three different strategies. We show the R-DSC of 6.25%, 25%, 50% annotated slices for Task A, and 12.5%, 25%, 50% for Task B.

We can see that our strategy is better than the other three strategies from Fig. 5 and Table 2. By using 15% annotated slices, our framework attains comparative performance as full annotation in Task1, and 30% annotated in Task2. The specific segmentation results are shown in Fig. 6.

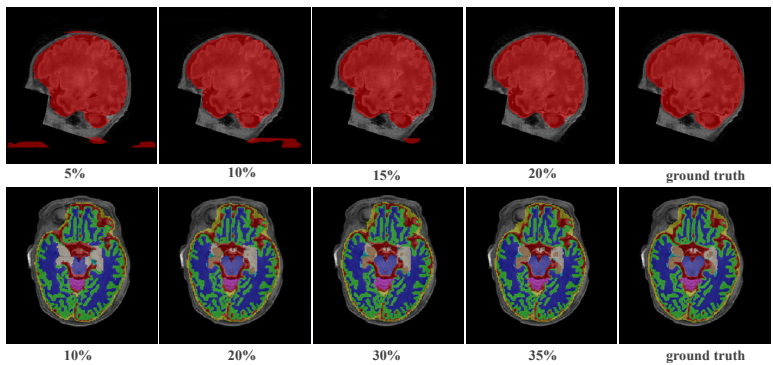

**Fig. 6.** This figure shows the improvement of segmentation results as the proposed framework proceeds. The percentage denotes the annotation ratio.

**Table 2.** Comparison with three different strategies using sparsely annotated training sets

| Strategy-ratio | F1 score (Task1) | | F1 score (Task2) | |
|---|---|---|---|---|
| | Part A | Part B | Part A | Part B |
| Random query-50% | 0.975 | 0.970 | 0.846 | 0.846 |
| Equal-interval query-50% | 0.982 | 0.975 | 0.859 | 0.854 |
| Uncertainty query-50% | 0.983 | 0.977 | 0.872 | 0.862 |
| Attention-guided AL-**15%** | **0.984** | **0.979** | 0.857 | 0.855 |
| Attention-guided AL-**30%** | 0.984 | 0.979 | **0.882** | **0.872** |

## 4     Conclusion

In this paper, we propose a sparse annotation strategy based on attention-guided active learning. Our framework provides four main contributions: (1) Attention-embedded 3D u-net improves segmentation performance remarkably. (2) Two metrics show high correlation with segmentation accuracy, and can be used to estimate the segmentation quality of samples without ground truth. (3) Attention-guided uncertainty measure is a brand new attempt in active learning, which may provide a direction for similar work. (4) Our proposed framework achieves the comparative results as full annotation with less labeling cost, greatly reducing the annotation time.